\documentclass[letterpaper, 10 pt, conference]{ieeeconf}  

\IEEEoverridecommandlockouts                              

\usepackage{amsmath,amsfonts}
\usepackage{algorithmic}
\usepackage{array}
\usepackage{xcolor}
\usepackage{soul} 
\usepackage[caption=false,font=normalsize,labelfont=sf,textfont=sf]{subfig}
\usepackage{textcomp}
\usepackage{stfloats}
\usepackage{url}
\usepackage{verbatim}
\usepackage{graphicx}
\usepackage{cite}

\usepackage[linesnumbered,ruled]{algorithm2e}
\SetKwBlock{Init}{Initialize}{end}
\SetKwBlock{Train}{Train Model}{end}
\SetKwBlock{Extract}{Extract Action Effect Features}{end}
\SetKwBlock{Plan}{Plan Generation}{end}
\SetKwBlock{Parallel}{Parallel Executation}{end}

\usepackage{booktabs}
\usepackage{makecell}
\usepackage{multirow}
\usepackage{threeparttable}

\usepackage[hang,flushmargin]{footmisc}
\makeatletter
\newcommand{\algorithmfootnote}[2][\footnotesize]{%
	\let\old@algocf@finish\@algocf@finish
	\def\@algocf@finish{\old@algocf@finish
		\leavevmode\rlap{\begin{minipage}{\linewidth}
				#1#2
		\end{minipage}}%
	}%
}
\makeatother 

\usepackage{xcolor}
\definecolor{c1}{HTML}{177cb0}
\definecolor{c2}{HTML}{065279}

\usepackage{hyperref}
\hypersetup{
	pagebackref=false,
	hypertex=true,
	colorlinks=true,
	linkcolor=blue,
	anchorcolor=black,
	citecolor=blue,
	urlcolor=magenta
}

\usepackage{bm}

\title{\LARGE \bf
FLAM: \underline{F}oundation Model-Based Body Stabilization for Humanoid \underline{L}ocomotion \underline{a}nd \underline{M}anipulation
}

\author{Xianqi Zhang$^{1}$, Hongliang Wei$^{1}$, Wenrui Wang$^{1}$, Xingtao Wang$^{1}$, Xiaopeng Fan$^{1, 2}$, Debin Zhao$^{1, 2}$
	\thanks{
			$^{1}$ X. Zhang, H. Wei, W. Wang, X, Wang, X. Fan, and D. Zhao are with the Faculty of Computing, Harbin Institute of Technology, Harbin 150001, China.
	}%
	\thanks{
			$^{2}$ X. Fan, and D. Zhao are also with the Peng Cheng Laboratory, Shenzhen 518000, China.
	}
	\thanks{
		 Email: {\{zhangxianqi, hlwei, 21B303001\}@stu.hit.edu.cn, \{xtwang, fxp, dbzhao\}@hit.edu.cn}
	}
	\thanks{
		Corresponding author: Xingtao Wang.
	}
}

\begin{document}

\maketitle
\thispagestyle{empty}
\pagestyle{empty}

\begin{abstract}
Humanoid robots have attracted significant attention in recent years. 
Reinforcement Learning (RL) is one of the main ways to control the whole body of humanoid robots.
RL enables agents to complete tasks by learning from environment interactions, guided by task rewards.
However, existing RL methods rarely explicitly consider the impact of body stability on humanoid locomotion and manipulation.
Achieving high performance in whole-body control remains a challenge for RL methods that rely solely on task rewards.
In this paper, we propose a Foundation model-based method for humanoid Locomotion And Manipulation (FLAM for short).
FLAM integrates a stabilizing reward function with a basic policy.
The stabilizing reward function is designed to encourage the robot to learn stable postures, thereby accelerating the learning process and facilitating task completion.
Specifically, the robot pose is first mapped to the 3D virtual human model.
Then, the human pose is stabilized and reconstructed through a human motion reconstruction model. 
Finally, the pose before and after reconstruction is used to compute the stabilizing reward.
By combining this stabilizing reward with the task reward, FLAM effectively guides policy learning.
Experimental results on a humanoid robot benchmark demonstrate that FLAM outperforms state-of-the-art RL methods, highlighting its effectiveness in improving stability and overall performance.
Project page: \url{https://xianqi-zhang.github.io/FLAM}
\end{abstract}

\section{Introduction}
\label{section_1}

Humanoid robots are expected to replace humans in heavy, repetitive, and dangerous tasks, and have developed rapidly in recent years.
Humanoid robots involve multiple research fields, such as perception, human-computer interaction, planning and control, etc. 
To promote its development, many hardware devices \cite{ficht2021bipedal, cao2024ai, tong2024advancements} and simulation environments \cite{humanoidbench, chernyadev2024bigym, wang2024grutopia} have been proposed in recent years.
Compared with earlier types of robots, such as arms and quadrupeds, humanoid robots are significantly more complex and challenging to control. 
On one hand, they possess a greater number of Degrees of Freedom (DoF), making their movements more intricate. 
On the other hand, they are expected to exhibit human-level intelligence, which further complicates the study of related planning and control methods.

For humanoid robotics tasks, locomotion \cite{bao2024deep} and manipulation \cite{gams2022manipulation, kroemer2021review} are the two main types.
1) 
Locomotion capabilities aim to let robots navigate in complex environments.
Although bipedal and humanoid robots are more human-like and therefore better suited for human surroundings, they also face many challenges,
such as balance and stability, complex terrain navigation, human-like motion, etc.
Radosavovic \textit{et al.} \cite{radosavovic2024humanoid} consider humanoid control as a next token prediction task and propose a causal transformer-based method for real-world locomotion.
Ding \textit{et al.} \cite{ding2024robust} propose a sequential convex optimization approach based on the inverted pendulum model and achieves good performance.
2) 
Manipulation capabilities aim to enable robots to interact with objects in a reasonable manner, which is an essential requirement for high-level intelligent robots.
To complete manipulation tasks, it is necessary to solve difficulties such as perception, motion and force control, human-like grasp, etc.
Rakita \textit{et al.} \cite{rakita2019shared} propose a shared-control method, including a bimanual action vocabulary constructed by analyzing how people perform two-hand manipulations.
Grannen \textit{et al.} \cite{grannen2023stabilize} simplifies bimanual manipulation tasks by separating the roles of two arms, with the stabilizing arm keeping the object in place while the acting arm performs the task.

To control the whole body of robots, Reinforcement Learning (RL) has gained widespread attention over the years \cite{bao2024deep, han2023survey}.
Garc{\'\i}a \textit{et al.} \cite{garcia2020teaching} teach a humanoid robot how to walk with safe RL.
Haarnoja1 \textit{et al.} \cite{haarnoja2024learning} leverage deep RL to enable a humanoid robot to acquire movement skills and play soccer.
However, previous RL methods rarely explicitly consider the impact of body stability on humanoid locomotion and manipulation tasks.
Additionally, the high degrees of freedom (DoF) in humanoid robots, coupled with the challenge of designing effective reward functions, make achieving high performance remains challenging for RL methods that rely solely on task rewards.

Recently, an increasing number of studies have leveraged findings from other research fields to advance robot development, with the goal of enhancing robot intelligence.
On the one hand,
foundation models \cite{hu2023toward, firoozi2023foundation} are becoming increasingly common in robotics research,
such as Large Language Models (LLMs) \cite{zeng2023large},
Vision-Language Model (VLM) \cite{zhang2024vision},
Contrastive Language Image Pretraining (CLIP) encoders \cite{radford2021learning, khandelwal2022simple},
Segment Anything Model (SAM) \cite{kirillov2023segment},
etc.
On the other hand,
3D virtual human models (i.e., SMPL \cite{SMPL} and SMPL-X \cite{SMPL-X}) and related human motion datasets (e.g., AMASS \cite{mahmood2019amass}) have been widely used to assist policy learning in achieving human-like behaviors. 
Motion data are often treated as motion priors or expert demonstrations for humanoid robots \cite{he2024omnih2o, dugar2024learning, ji2024exbody2} and simplified skeletal representations like matchstick men \cite{luo2024grasping, luo2023perpetual, luo2024smplolympics}.

For a person, the difficulty of the tasks being learned typically increases gradually. 
A fundamental and crucial step is learning to stand and maintain a stable posture, which serves as the foundation before progressing to more complex tasks such as locomotion and manipulation.
Inspired by this phenomenon and recent robotics research, 
in this paper, we propose a Foundation model-based body stabilization method for humanoid Locomotion And Manipulation (FLAM for short), which consists of a stabilizing reward function and a basic policy.
Specifically, a pre-trained foundation model, i.e., a human motion reconstruction model, is used to design the stabilizing reward function.
The reward encourages the robot to learn and maintain stable postures.
For a task, the stabilizing reward and the task reward are combined to guide the basic policy learning.

The main contributions of this paper are summarized as follows:
\begin{itemize}
	\item{
		A foundation model-based method named FLAM is proposed for humanoid locomotion and manipulation, which integrates a stabilizing reward function with a basic policy.
		Experimental results verify the effectiveness of the method.
	}
	\item{
		A stabilizing reward function is proposed, which is designed with a foundation model, i.e., a human motion reconstruction model. 
		The reward encourages the robot to learn and maintain stable postures, thereby accelerating the policy learning process and facilitating task completion.
	}
\end{itemize}

The rest of this paper is organized as follows. 
Related works are briefly reviewed in Section \ref{section_2}. 
The proposed framework is described in Section \ref{section_3}. 
Experimental results are presented in Section \ref{section_4}. 
Section \ref{section_5} provides the conclusion.

\section{Related Works}
\label{section_2}
In this section, we review the related works of humanoid locomotion and manipulation, whole-body control and reinforcement learning, and foundation models for robotics, respectively.

\subsection{Humanoid Locomotion and Manipulation}
Humanoid locomotion and manipulation, as two primary types of robot tasks, have been extensively studied for a long time \cite{gu2025humanoid},
with the goal of enhancing robot robustness, efficiency, and adaptability across diverse environments and tasks.

Locomotion capabilities allow robots to move through complex environments. 
Cui \textit{et al.} \cite{cuiadapting} introduce a two-phase training framework with RL for real-world humanoid locomotion.
Dantec \textit{et al.} \cite{dantec2022whole} propose a whole-body model predictive control scheme based on differential dynamic programming, achieving promising results.
Gao \textit{et al.} \cite{gao2022invariant} use an invariant extended Kalman filter to estimate robot pose and velocity for locomotion.
Although humanoid and bipedal designs make them more compatible with human-centric environments, they also introduce challenges like maintaining balance, navigating rough terrain, and achieving natural movement.
These challenges make robust locomotion for humanoid robots still require further research.

Manipulation capabilities are crucial for robots to interact effectively with objects, making them a fundamental requirement for advanced intelligent systems.
Successful manipulation requires overcoming challenges such as object perception, motion and force control, and human-like grasping.
Moreover, humanoid robots typically have two arms and dexterous hands, further increasing the complexity of manipulation tasks.
Zhao \textit{et al.} \cite{zhao2023learning} present a low-cost system that performs bimanual manipulation.
Ze \textit{et al.} \cite{ze2024generalizable} propose a real-world robotic system that collects data via teleoperation and adopts a 3D diffusion policy for knowledge learning.
Although existing methods have achieved promising results in specific scenarios, developing general grasping skills remains a challenge due to the vast diversity of objects and the high DoF of dexterous hands.

\subsection{Whole-Body Control and Reinforcement Learning}
Whole-Body Control (WBC) \cite{moro2019whole, ficht2020fast, cheng2024expressive} is a crucial aspect of robotics, as it enables robots to seamlessly coordinate locomotion and manipulation. 
To achieve this, control methods must effectively orchestrate multiple DoF to complete complex tasks while ensuring safety and stability.
Traditional WBC approaches often formulate control as an optimization problem, maintaining balance and avoiding collisions, while satisfying various constraints, including joint limits, friction, and actuation limits. 
Recently, Reinforcement Learning (RL) has been widely used to enhance whole-body coordination, providing robots with better adaptability and performance in complex scenarios \cite{lu2024mobile, cheng2024expressive, zhuang2024humanoid}.

RL refers to learning from interactions with the environment, which typically defines a suitable reward function based on the task and uses it to guide the learning of the policy. 
The policy can be either value-based, which estimates action values, or policy-based, which learns a policy directly. 
Both approaches enable the learning process to gradually produce appropriate actions to complete the task.
RL methods can also be categorized into model-based \cite{moerland2023model, luo2024survey, tdmpc2, dreamerv3} and model-free \cite{ccalicsir2019model, ppo, sac, cqn, cqn-as}, each with distinct characteristics and applications.
1) 
Model-based RL uses a model of the environment to predict future states and rewards, allowing the agent to estimate what will happen as a result of its actions in the current state.
By simulating possible outcomes of an action, the agent can plan more efficiently.
This approach is often sample-efficient and enables long-term decision-making but can be computationally expensive and reliant on the accuracy of the environment model.
2) 
Model-free RL, in contrast, does not rely on an explicit environment model.
Instead, the agent learns optimal policies directly through trial and error.
Model-free RL methods are usually more flexible and well-suited for high-dimensional environments, but they often require large amounts of interaction data and struggle with sample inefficiency.

\subsection{Foundation Models for Robotics}
Foundation models have emerged as a transformative approach in robotics, leveraging large-scale pre-trained neural networks to improve perception, decision-making, and generalization \cite{hu2023toward, firoozi2023foundation}. 
Inspired by the success of foundation models in natural language processing (e.g., LLMs \cite{zeng2023large}) and computer vision (e.g., VLM \cite{zhang2024vision}, CLIP \cite{radford2021learning, khandelwal2022simple}, SAM \cite{kirillov2023segment}), researchers have started applying similar architectures to robotic planning and control.
For example, 
Huang \textit{et al.} \cite{huang2022language} use LLM as zero-shot planners for task planning.
Stone \textit{et al.} \cite{stone2023open} leverage VLM for open-world object manipulation.
Huang \textit{et al.} \cite{huang2023voxposer} propose VoxPoser for robotic manipulation, which combines LLM, VLM, and SAM.

Recently, to enable humanoid robots to learn human-like behaviors, 3D virtual human models (i.e., SMPL \cite{SMPL} and SMPL-X \cite{SMPL-X}) and related human motion datasets (e.g., AMASS \cite{mahmood2019amass}) are used in many works.
For example,
He \textit{et al.} \cite{he2024omnih2o} train a policy to imitate human motions in AMASS.
Dugar \textit{et al.} \cite{dugar2024learning} propose the Masked Humanoid Controller (MHC), which is also learned from AMASS.

These methods typically rely on human motion datasets as a motion prior or expert demonstrations to train a policy for task completion.
In contrast, our approach utilizes SMPL-X \cite{SMPL-X} for expression and leverages a foundation model for reward function design. 
This allows us to assess posture stability directly through the pre-trained foundation model. 
Since our goal is not to imitate human behaviors, the human motion dataset is unnecessary.

\section{Framework}
\label{section_3}
\begin{figure}[!t]
	\centering
	\includegraphics[width=3.3in]{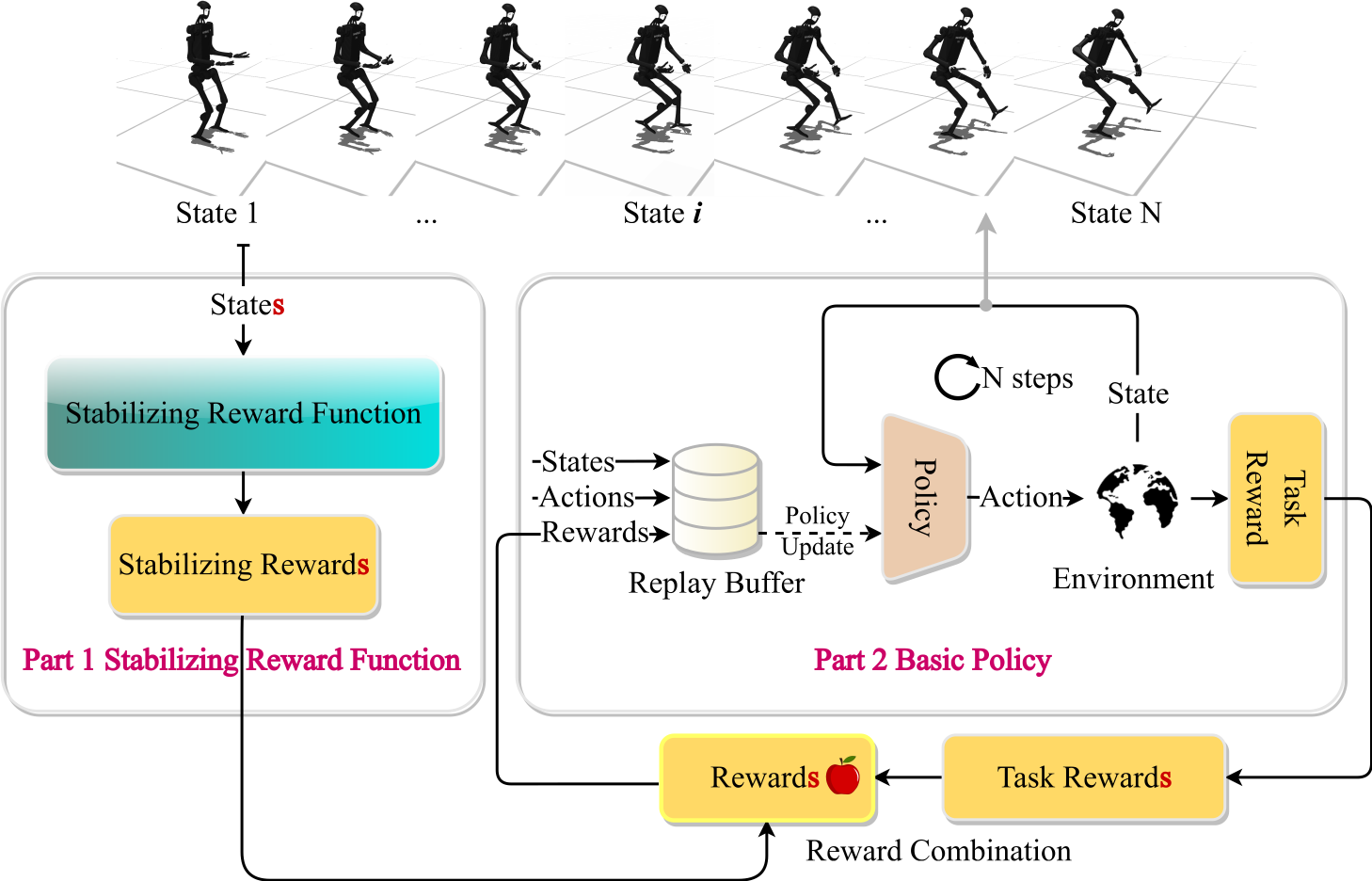}
	\caption{
		The framework of FLAM.
	}
	\label{fig_pipeline}
\end{figure}

The proposed method FLAM contains two parts.
First, the stabilizing reward function is proposed with the human motion reconstruction foundation model, which is used to enable the robot to learn stable poses.
Then, a policy is trained to complete tasks according to the stabilizing reward and the task reward.
The framework is shown in Fig.~\ref{fig_pipeline}.
In the following subsections, we will elaborate on the two parts, respectively.

\subsection{Stabilizing Reward Function}
\label{section_3_1}

\begin{figure}[t]
	\centering
	\includegraphics[width=3.2in]{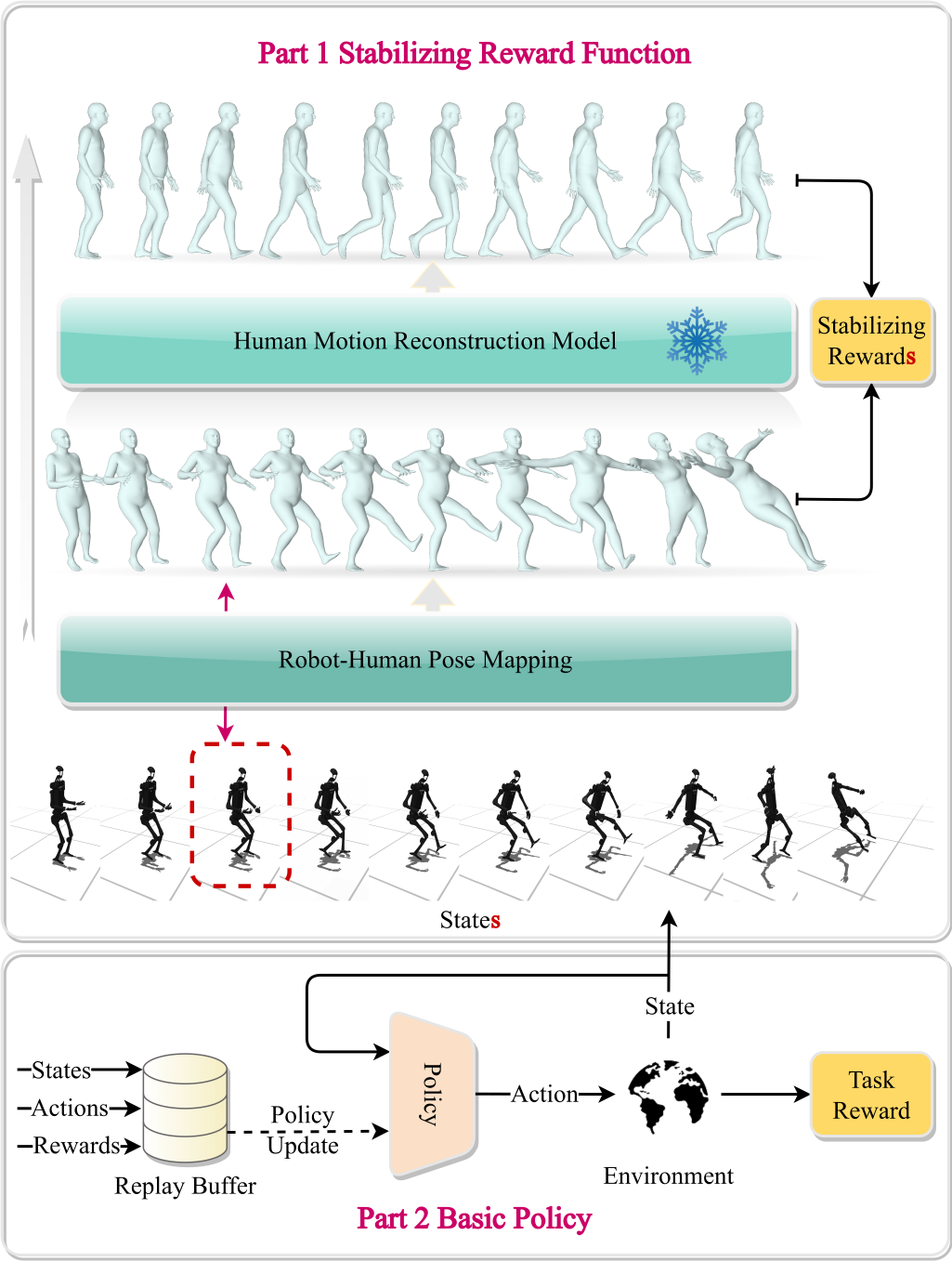}
	\caption{The overview of the stabilizing reward function.
	}
	\label{fig_stabilize}
\end{figure}

The stabilizing reward function is proposed to help the robot learn stable poses, thereby accelerating the learning process and facilitating task completion.
The input of the function is a trajectory segment, consisting of robot body states.
The output is also a trajectory segment, but it consists of stable human poses, expressed with SMPL-X \cite{SMPL-X}.
The overview is shown in Fig.~\ref{fig_stabilize}.

The processing flow of the method mainly involves three steps. 
First, the trajectory segment composed of robot poses/states is mapped to the trajectory composed of human poses. 
Second, the human trajectory is processed by a human motion reconstruction foundation model to produce a stable motion trajectory. 
Finally, the reward is calculated based on the original robot trajectory and the stable human trajectory.
We will now introduce each step in detail.

\subsubsection{Robot-human pose mapping}
\begin{figure}[t]
	\centering
	\includegraphics[width=2.9in]{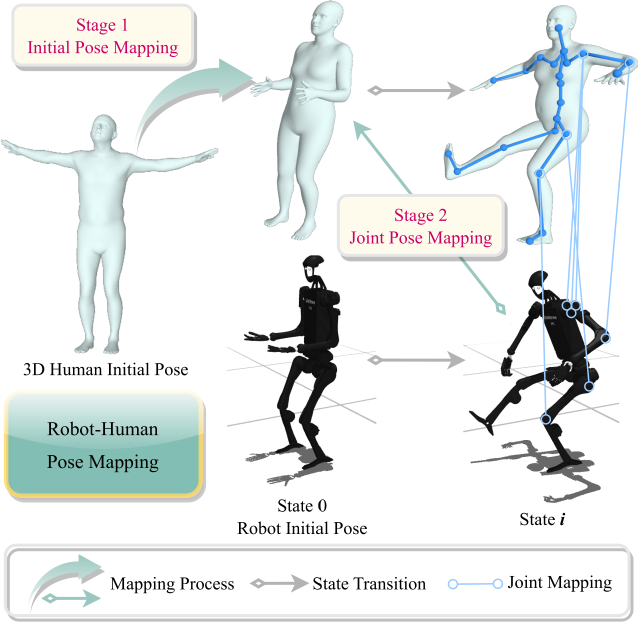}
	\caption{
		The robot-human pose mapping process.
		Joint mappings are simplified for clarity.
	}
	\label{fig_mapping}
\end{figure}

To map the two poses, the human pose is first initialized to match the initial pose of the robot. 
In other words, their initial zero pose are aligned.
Then, for a specific state, each human joint is adjusted from the aligned initial pose, with its rotation angle determined by the relative orientation of the corresponding robot joint.
The joints of the two are usually not completely matched: 
a) In some anatomical areas, the human has more joints than the humanoid robot. 
Certain human joints are selected for mapping, while redundant joints remain at the initial pose. 
For example, one of the human spine joints is mapped to the waist joint of the robot.
b) In other anatomical areas, the human has fewer joints. 
Different coordinate axes of a human spherical joint will be mapped to three revolute joints of the humanoid robot, such as shoulder joints.
The robot-human pose mapping can be expressed as follows, and the process is shown in Fig.~\ref{fig_mapping}.

\begin{equation}\label{eq1}
	\begin{aligned}
		\bm s_h  =   {\rm \mathcal{F}_{M}}(\bm s_r), \\
	\end{aligned}
\end{equation}
where $\bm s_r$ is the robot pose and $\bm s_h$ is the aligned human pose.

\subsubsection{Human motion reconstruction}
To obtain stable poses, a diffusion-based foundation model, RoHM \cite{zhang2024rohm}, is employed for human motion stabilization and reconstruction.
Since the model needs to compute information such as the movement path, both the input and output are trajectory segments.
Specifically, the input of the model is noisy human poses, and the output is stable human poses.
The reconstruction operation can be expressed as:

\begin{equation}\label{eq2}
	\begin{aligned}
		\hat{\mathbb{S}}  =  {\rm \mathcal{F}_{R}}(\mathbb{S}). \\
	\end{aligned}
\end{equation}
Here, ${\rm \mathcal{F}_{R}}$ is the foundation model used to reconstruct human motion.
$\mathbb{S}$ and $\hat{\mathbb{S}}$ are trajectory segments, i.e., 
$\mathbb{S} = \{\bm s_h^0, \bm s_h^1, ..., \bm s_h^{l_s}\}$
and 
$\hat{\mathbb{S}} = \{\hat{\bm s}_h^0, \hat{\bm s}_h^1, ..., \hat{\bm s}_h^{l_s}\}$, 
$l_s$ is the segment length.

\subsubsection{Reward function design}
The reward is calculated based on the difference between the aligned (noisy) human pose and the reconstructed (stable) human pose.
When the poses before and after reconstruction are similar, it indicates that the current robot pose is stable, triggering a reward.
Pose similarity is evaluated by comparing corresponding joints. 
A joint is considered similar if the difference in its position and orientation is below a specified threshold. 
If the total number of similar joints surpasses a certain value, the two poses are classified as similar.
The similarity function is expressed as follows:

\begin{equation}\label{eq3}
	\begin{aligned}
		{\rm \mathcal{F}_{S}}(\bm s_h, \hat{\bm s}_h)  =  \sum_{i=0}^{N} {\rm  \Gamma}(\bm s_h^i, \hat{\bm s}_h^i) \\
	\end{aligned}
\end{equation}
with
\begin{equation}\label{eq4}
	\begin{aligned}
		{\rm  \Gamma}(\bm s_h^i, {\hat{\bm s}_h^i})  = 
		\left\{
		\begin{array}{lr}
			\!\! r_j, \quad if \:||\bm s_h^i - \hat{\bm s}_h^i||_2^2 \leq t_j\\
			\!\! 0, \quad \: \: Others.
		\end{array}
		\right.\\
	\end{aligned}
\end{equation}
Here, $\bm s_h$ represents the aligned human pose, and $\hat{\bm s}_h$ denotes the reconstructed human pose.
$N$ is the number of joints.
$\bm s_h^i$ corresponds to the $i\mbox{-}th$ joint pose.
$r_j$ is the joint reward.
$t_j$ is the joint distance threshold.

The stabilizing reward function can be formulated as

\begin{equation}\label{eq5}
	\begin{aligned}
		{\rm R_S}\left(\bm s_r\right)  = 
		\left\{
		\begin{array}{lr}
			\!\! {\rm \mathcal{F}_{S}}(\bm s_h, \hat{\bm s}_h), \quad           if \: {\rm \mathcal{F}_{S}}(\bm s_h, \hat{\bm s}_h) \geq t_s\\
			\!\! 0, \qquad \qquad \quad \:     Others,
		\end{array}
		\right.\\
	\end{aligned}
\end{equation}
where $t_s \! = \! \bar{N} \cdot r_j$ is the similarity threshold, $ \bar{N}$ is the number of expected similar joints.
$\bm s_h$ and $\hat{\bm s}_h$ are calculated with Eq.(\ref{eq1}) and Eq.(\ref{eq2}), respectively.

\subsection{Basic Policy}
\label{section_3_3}

\subsubsection{Architecture}
Since policy architecture is not the focus, a model-based policy, specifically TD-MPC2\cite{tdmpc2}, is adopted directly.

The basic policy, i.e., TD-MPC2\cite{tdmpc2}, is a data-driven Model Predictive Control (MPC) framework that uses Temporal Difference (TD) learning to train a latent dynamics model and a value function. 
The latent dynamics model is used to optimize short-term rewards, and the value function is used to estimate long-term returns.
The policy consists of five components, which can be expressed as follows.
More details can be found in \cite{tdmpc2}.

\begin{equation}
	\label{eq6}
	\begin{array}{ll}
		\text{Encoder} & \bm z = {\rm E}(\bm s) \\
		\text{Latent dynamics} & {\bm z}_{t+1} = {\rm D}({\bm z}_t, \bm a) \\
		\text{Reward} & \hat{r} = {\rm R}(\bm z, \bm a) \\
		\text{Terminal value} & \hat{q} = {\rm Q}(\bm z, \bm a) \\
		\text{Policy prior} & \hat{\bm a} = {\rm P}(\bm z),
	\end{array}
\end{equation}
where $\bm s$ donates the state, $\bm a$ represents the action, and $\bm z$ is the latent representation.

\subsubsection{Training Strategy}
The policy is trained according to the stabilizing reward and the task reward.
A straightforward method is adopted for reward combination, which can be described as follows:

\begin{equation}\label{eq7}
	\begin{aligned}
		{\rm R} = {\rm R_T} + \lambda \frac{q}{l_e} \cdot {\rm R_S}.
	\end{aligned}
\end{equation}
Here, $\rm R_T$ is the task reward function.
$\lambda$ is the scaling factor, which reflects the importance of stabilizing reward relative to task-specific reward.
A small value suggests that stability is less crucial, as in tasks like \textit{Crawl} and \textit{Sit}.
In contrast, a large value indicates that stability plays a more significant role.
$q$ is the expected return, which is related to the specific task.
$l_e$ is the max episode length, i.e., 1000 in the experiments.
$q/{l_e}$ is used to estimate the expected contribution of each step.
$\rm R_T$ is typically used to encourage the policy to generate actions that positively contribute to task completion, with a higher reward assigned upon task success. 
A simple approach is to assign a nonzero reward only upon task completion, known as the sparse reward setting.
In contrast, the dense reward setting provides feedback throughout the process by rewarding behaviors that contribute to task completion.
Since an existing benchmark is used for experiments, $\rm R_T$ is predefined by the benchmark.

The policy update process differs slightly from previous works, as the stabilizing reward function takes trajectory segments as input. 
The process consists of the following steps:
a) The policy interacts with the environment for $l_s$ steps, generating a trajectory segment of $l_s$ states.
b) Stabilizing rewards are computed based on the collected trajectory segment.
c) Final rewards are calculated using Eq.(\ref{eq7}) and stored in the replay buffer.
d) The policy is updated by sampling batch data from the replay buffer, repeating sampling and update $l_s$ times.

\begin{figure*}[!t]
	\centering
	\includegraphics[width=.97\linewidth]{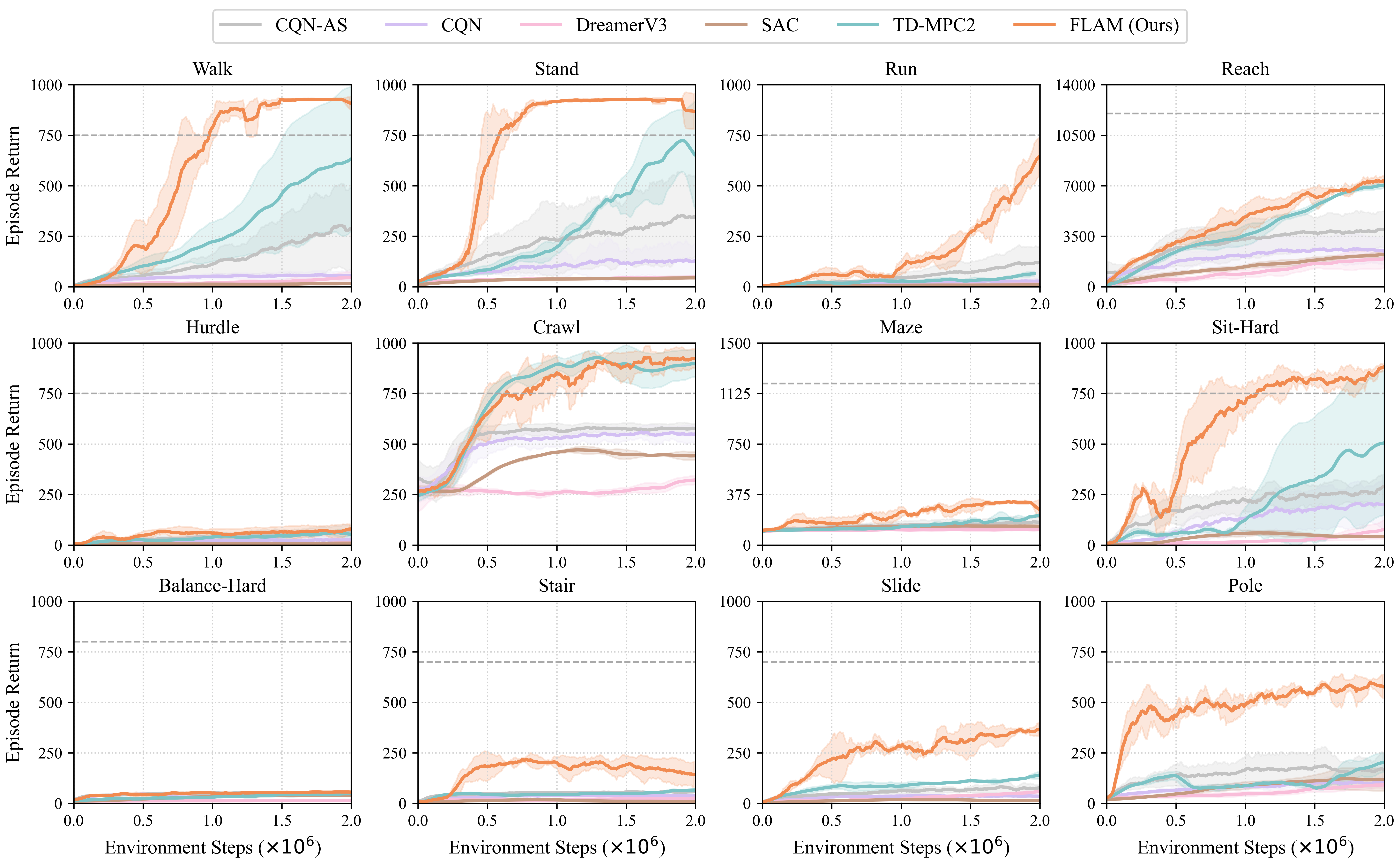}
	\caption{
		Performance comparison of methods on locomotion tasks.
		The dashed lines qualitatively indicate task success. 
	}
	\label{fig_locomotion_results}
\end{figure*}

\begin{table}[t]
	\centering
	\footnotesize
	\renewcommand{\arraystretch}{1.6}
	\setlength\aboverulesep{0pt}\setlength\belowrulesep{0pt}
	\tabcolsep=12.0pt
	\caption{Hyperparameter Settings}
	\label{tab_param}
	\begin{threeparttable}
		\begin{tabular}{ c | c | c }
			\toprule
			Equation & Parameter & Value \\
			\midrule
			Eq.(\ref{eq2}) & segment length $l_s$        & 145              \\
			Eq.(\ref{eq4}) & joint reward $r_j$          & 0.1              \\
			Eq.(\ref{eq4}) & joint threshold $t_j$       & 0.1              \\
			Eq.(\ref{eq5}) & similarity threshold $t_s$  & 1.5              \\
			Eq.(\ref{eq7}) & scaling factor $\lambda$    & [0.5, 1.0] \\
			Eq.(\ref{eq7}) & expected return $q$         & [350, 750]       \\
			\bottomrule
		\end{tabular}
	\end{threeparttable}
\end{table}

\section{Experiments}
\label{section_4}
In this section, extensive experiments are carried out to compare FLAM with state-of-the-art methods.
First, the experimental setup is introduced. 
Second, performance comparison results are provided.
After that, some ablation experiments are presented. 
Finally, limitations and future works are provided.

\subsection{Experimental Setup}

\subsubsection{Benchmark}
The simulated humanoid robot benchmark Humanoid-Bench \cite{humanoidbench} is adopted for the experiments.
The robot is a Unitree H1, equipped with dexterous hands, with a total of 61 DoF (21 DoF for the body and 40 DoF for the two hands).
The benchmark uses a dense reward setting.

\subsubsection{Evaluation Metrics}
The return value, defined as the sum of rewards over the trajectory, is used to evaluate the performance of the methods.

\subsubsection{Baselines}
We compare the proposed method FLAM with 
model-based RL methods:
TD-MPC2 \cite{tdmpc2} and
DreamerV3 \cite{dreamerv3},
model-free RL methods:
CQN \cite{cqn},
CQN-AS \cite{cqn-as},
SAC \cite{sac}.

\subsubsection{Implementation Details}
Hyperparameter settings of FLAM are listed in Table \ref{tab_param}.
The scaling factor $\lambda$ is typically set to $1.0$ for most tasks. 
However, for tasks such as \textit{Crawl} and \textit{Sit}, which do not require high body stability, a lower value of $0.5$ is used.
For the expected return value $q$ in Eq.(\ref{eq7}), since it is similar for some tasks, we fix it to $750$ for locomotion and $350$ for manipulation to simplify implementation.
In addition, it is important to note that the architecture, number of parameters, and settings of the basic policy are the same as that in \cite{humanoidbench, tdmpc2}.


Baseline results are mainly taken from Humanoid-Bench and CQN-AS.
The results of CQN and CQN-AS on \textit{Sit-Hard, Balance-Hard, Stair, Slide}, and \textit{Pole}, are reproduced according to the official implementation of CAN-AS.

\begin{figure*}[!t]
	\centering
	\includegraphics[width=.97\linewidth]{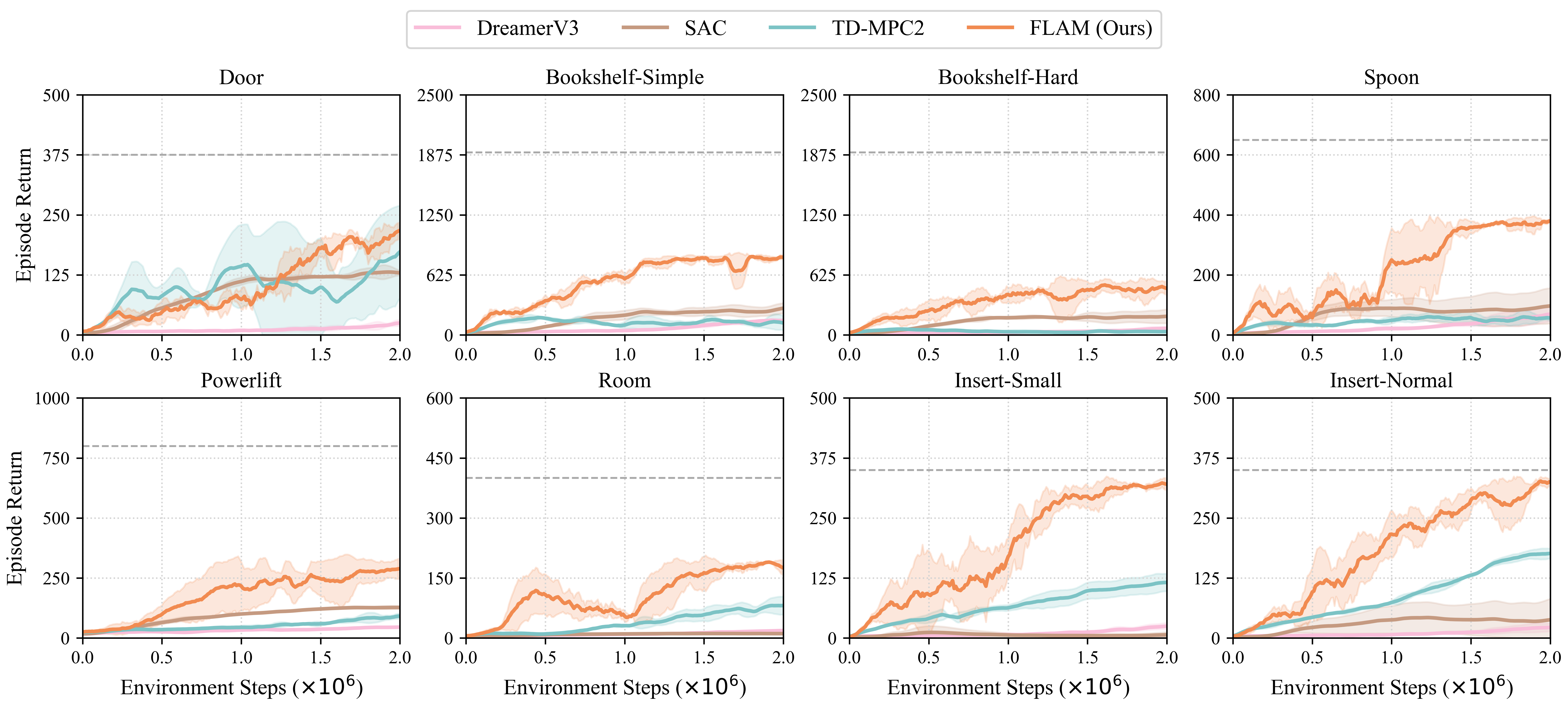}
	\caption{
		Performance comparison of methods on manipulation tasks.
		The dashed lines qualitatively indicate task success. 
	}
	\label{fig_manipulation_results}
\end{figure*}

\begin{figure*}[t]
	\centering
	\includegraphics[width=.96\linewidth]{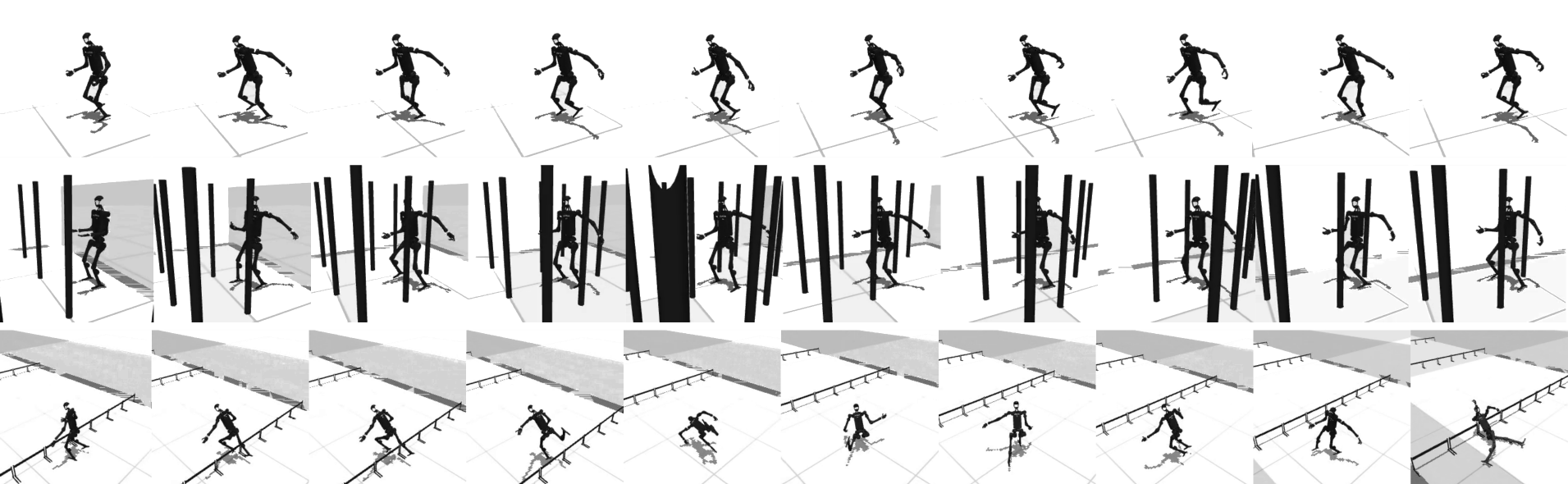}
	\caption{
		Qualitative results. 
		The visualization of FLAM on successful tasks (\textit{Run, Pole}) and a failed task (\textit{Hurdle}).
	}
	\label{fig_qualitative_results}
\end{figure*}

\subsection{Performance Comparison}
	
Experimental results are reported over 2 million steps, as shown in Fig.~\ref{fig_locomotion_results} and Fig.~\ref{fig_manipulation_results}.

\subsubsection{Locomotion}
The performance comparison of the methods on locomotion tasks is shown in Fig.~\ref{fig_locomotion_results}.
FLAM achieves state-of-the-art performance on various tasks, such as \textit{Walk, Stand, Run, Sit-Hard}, and \textit{Pole}, highlighting the importance of body stability in humanoid robot locomotion.
For tasks like \textit{Maze, Stair}, and \textit{Slide}, FLAM yields small improvements, 
while for \textit{Reach, Crawl, Hurdle}, and \textit{Balance-Hard}, its impact is negligible.

We analyze the reasons as follows: 
1) The training data for the foundation model primarily consists of human motions in planar environments. As a result, its stability estimation becomes inaccurate in non-planar environments, such as \textit{Balance-Hard, Stair} and \textit{Slide}. 
2) The proposed method does not address the stability recovery problem, specifically, how to restore the robot from an unstable state to a stable state during movement, which is crucial for tasks like \textit{Hurdle}.
3) The inherent challenges of RL methods, such as environment exploration, remain a bottleneck for certain tasks, e.g., \textit{Maze}.

\subsubsection{Manipulation}
The performance comparison of the methods on manipulation tasks is shown in Fig.~\ref{fig_manipulation_results}.
FLAM outperforms baselines on all tasks, demonstrating its effectiveness.
However, none of the methods successfully completed the tasks. The primary challenge lies in controlling the dexterous hands with high DoF, which is difficult even with a stable body.

As shown in Fig.~\ref{fig_qualitative_results}, we provide qualitative results, specifically, the visualization of FLAM on successful tasks (\textit{Run, Pole}) and a failed task (\textit{Hurdle}).

\subsection{Ablation Experiments}
FLAM, which consists of a stabilizing reward function and a basic policy, simplifies to the basic policy TD-MPC2 when the stabilizing reward is entirely removed. 
Therefore, the effect of the proposed reward function can be analyzed by comparing FLAM with TD-MPC2 in Fig.~\ref{fig_locomotion_results} and Fig.~\ref{fig_manipulation_results}.

\begin{figure*}[t]
	\centering
	\includegraphics[width=.97\linewidth]{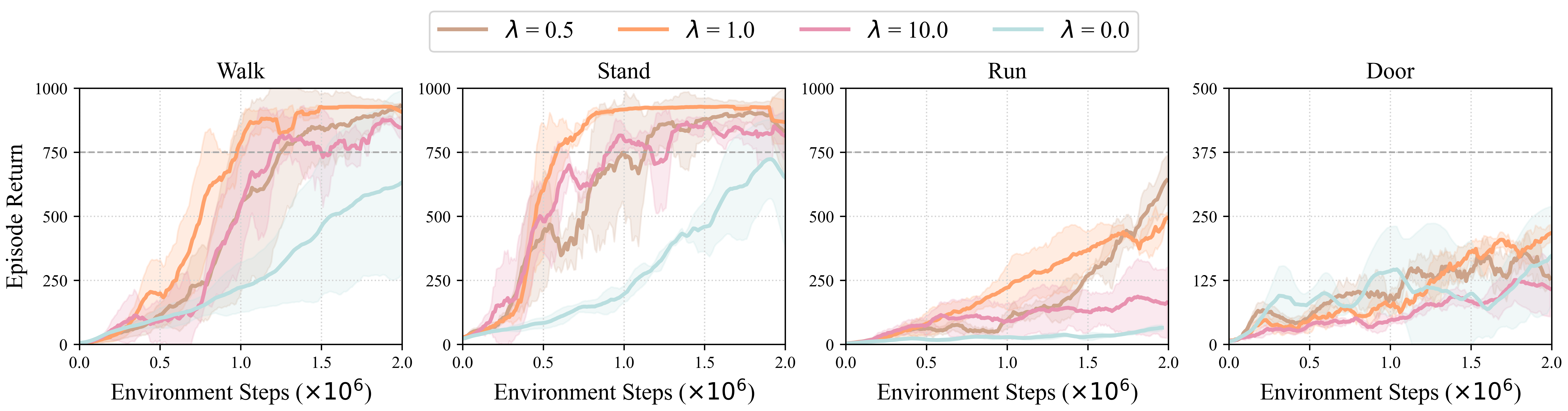}
	\caption{
		Ablation study of the scaling factor $\lambda$.
	}
	\label{fig_ablation_lambda}
\end{figure*}

We also perform ablation experiments on the scaling factor $\lambda$ in the reward combination Eq.(\ref{eq7}).
The experiments are conducted on four tasks: three locomotion tasks (\textit{Walk, Stand, Run}) and a manipulation task (\textit{Door}).
As shown in Fig.~\ref{fig_ablation_lambda},
the tasks \textit{Walk}, \textit{Stand} and \textit{Door} are not highly sensitive to $\lambda$, but selecting suitable parameters can still enhance performance.
In contrast, for \textit{Run}, smaller $\lambda$ significantly improves performance.

It is important to set an appropriate scaling factor $\lambda$ for different tasks. 
If $\lambda$ is too small, the agent may struggle to develop the ability to maintain body stability. 
Conversely, if $\lambda$ is too large, the agent may over prioritize stability, merely imitating stable postures output by the foundation model while neglecting the task. 
An optimal value should allow the agent to maintain stability in a way that facilitates exploration, enabling it to gather task-related reward information and ultimately learn how to complete the task.

\subsection{Limitations and Future Works}

The proposed method FLAM shows promising performance, but still has many limitations.
1) Stability determination.
Since FLAM assesses the stability of the robot by comparing two distinct states, i.e., the states/poses before and after reconstruction, it tends to prioritize static or low-speed stability. 
However, for fast movements, it may be more appropriate to consider both preceding motions and predicted future states. 
For instance, a particular posture during running may appear unstable in isolation, yet it is essential for maintaining overall stability throughout the entire trajectory.
2) Stability restoration.
With the stabilizing reward, the policy can learn to maintain a stable posture while completing the task. 
However, determining whether the posture is stable is not enough to teach the policy how to regain stability after losing it, which is crucial in many tasks, such as \textit{Hurdle} and \textit{Slide}.
3) Reward combination.
While our approach considers the average task reward of each step in the reward combination, a more sophisticated dynamic fusion method may be more effective. 
Furthermore, it may be beneficial to explore alternative learning strategies, such as learning a stable posture before learning a task, or vice versa, rather than attempting to learn both simultaneously.
These directions will be investigated in the future.

\section{Conclusion}
\label{section_5}
In this paper, a foundation model-based method FLAM is proposed for humanoid locomotion and manipulation.
FLAM consists of a stabilizing reward function and a basic policy.
The reward function is designed with the human motion reconstruction foundation model, which is used to enable the robot to learn stable poses. 
The basic policy is trained to complete tasks according to the stabilizing reward and the task reward.
Experimental results on a humanoid benchmark have confirmed the superiority of the proposed method.




\bibliographystyle{IEEEtran}
\bibliography{./src/reference}
	
\end{document}